# The Problem of Coincidence in A Theory of Temporal Multiple Recurrence

**B.O. Akinkunmi**
(Department of Computer Science,
University of Ibadan, Ibadan, Nigeria
ope34648@yahoo.com)

# The Problem of Coincidence in A Theory of Temporal Multiple Recurrence

**Abstract**: Logical theories have been developed which have allowed temporal reasoning about eventualities (a la Galton) such as states, processes, actions, events, processes and complex eventualities such as sequences and recurrences of other eventualities. This paper presents the problem of coincidence within the framework of a first order logical theory formalizing temporal multiple recurrence of two sequences of fixed duration eventualities and presents a solution to it

The coincidence problem is described as: *if two complex eventualities (or eventuality sequences) consisting respectively of component eventualities $x_0, x_1,....,x_r$ and $y_0, y_1, ..,y_s$ both recur over an interval k and all eventualities are of fixed durations, is there a subinterval of k over which the incidence $x_t$ and $y_u$ for $0 \leq t \leq r$ and $0 \leq u \leq s$ coincide*. The solution presented here formalizes the intuition that a solution can be found by temporal projection over a cycle of the multiple recurrence of both sequences.

## 1. Introduction

Over time, a substantial body of literature has built up in the area of temporal knowledge representation and reasoning. Time has been studied as time points (or instants) and time intervals or moments. [2, 8, 10] Each of these views as to what should constitute the basic elements in the time ontology gives rise to different fundamental problems, and each proposition usually has as its motivation the solution to some particular reasoning problem or task.

In addition to the time ontology, there is also the need to identify the propositions whose truth values must be evaluated over time. A theory evaluating these is known as a theory of temporal incidence [17]. These, propositional entities, we refer to as *eventualities* in this paper following Galton[7] and generalizing Vila's fluent/state dichotomy[17]. Various taxonomies of temporal propositions have emerged over the years, starting from McDermott's dichotomy of facts and events to Allen's states/events/processes trichotomy [3], to Shoham's more elaborate classification, based on temporal properties [16].

Vila's survey [18] identified two basic tasks required to support the kind of temporal reasoning that needs to be done in medical diagnosis, planning, industrial process supervision and natural language understanding. These are *temporal consistency maintenance*, which involves maintaining consistency in a temporal data base while adding new entries, and *temporal question answering*, which involves "providing answers to queries" that require temporal knowledge.

In a sense temporal reasoning research has been addressing the problem of how to provide answers to the truth status (i.e. temporal incidence) of some propositional eventualities in the presence of certain well known temporal phenomenon such as recurrence. Koomen [10] had extended Allen's theory of action and time to accommodate the ability to reason about "recurrence". Precisely, Koomen developed a logical theory, which would enable temporal reasoners to reason about the concept of recurrence of both simple eventualities (such as events and states) and their sequences over time. Thus given that an alternating sequence of eventualities recurs over time, is it possible to find a future time interval within the interval of recurrence over which the latter of the two eventualities will be true. For example Koomen's theory enabled the reasoner to help a robot driver to reason on encountering a red light at a junction that the existence of a recurring sequence of red and green light will imply that at some point in the future the green light will come up. Thus encountering a red light at the junction is not good enough reason to abandon the driver's plan of getting to his destination. The temporal behaviour of recurrent (or repetitive) eventualities, has been studied in the literature[1] treating them as ontological entities.

Koomen's notion of recurrence contrasts with some others in the literature such as Pan[13]. While the notion of recurrence by Koomen implies a periodic repetition of an event, state or some other propositional eventuality with contiguous periods that a planner may need to track, Pan uses recurrence to refer to repeated incidence of a propositional eventuality at regular intervals, such as "a meeting holding every Tuesday" , as may arise in natural language discourse.

Our work here is motivated by the problem of how to enable a planner to cope with a slightly more complex scenario than Koomen's, which is reasoning about the possible incidence of the preconditions of an action in the context of regular *multiple* recurrence as opposed to a *single* one. Given that two sequences of eventualities $x_0, x_1,....,x_r$ and $y_0, y_1, ..,y_s$ (such that any pair of eventualities from the x sequence or the y sequence are mutually exclusive) both recur over some interval, and a planner requires the coincidence of an eventuality pair $x_p$ and $y_q$ from the sequences, in order to carry out a certain action. Can a planner infer the existence (or otherwise) of such an interval which is a subinterval of two intervals over which each of $x_p$ and $y_q$ are both true. In other words can a planner predict the incidence or otherwise of such a coincidence?

Suppose for example, that in addition to the traffic light at the junction in Koomen's example, there is a gate that is shut, at the top of every hour for fifteen minutes to enable some house keeping to be done in certain parts of the road network. Then, the planner must reason about the existence of some time in the future when both the gate will be open, and there will be green light at the junction.

Another simple example of this from a factory domain is a piece of factory machinery which must work continuously for five days which must be followed by a maintenance process that must last three continuous days, but the engineer will not be available on a Wednesday. In this case a scheduler must reason about the avoidance of a situation in which the maintenance period starts or ends on a Wednesday. In both cases the scheduler needs to acquire the capability to determine the possibility of the simultaneous incidence of two different conditions from two different recurring sequences.

In the first example, the planner must determine the existence of a future time interval over which the gate is open and there is green light at the junction i.e. the preconditions for his proposed action of driving past the point, while in the second, a scheduler must hope that he can avoid a situation in which one of the three consecutive maintenance days is a Wednesday. Now we consider the second example. If the factory starts working on a Monday, then the first maintenance days will be a Saturday to Monday; the next one will be a Sunday to Tuesday; while the third one will be a Monday to Wednesday. The next three maintenance sessions will all happen on a Wednesday. Thus our planner should know that it could not avoid a situation in which Wednesday is a maintenance day.

Our conclusion in the last paragraph was arrived at following an explicit temporal projection into the future, using our knowledge of the eventualities that make up the two recurring sequences, and their relative durations. In this case there are two sequences. The first one is the sequence of days of the week with eventualities <Monday, Tuesday, Wednesday…Sunday>, each with a duration of 1 day. The second sequence consists of two eventualities < Working_period, Maintenance_period>, where the first eventuality is of duration 5 days, while the other one is of duration three days.  This paper presents a logical way of formalizing the kind of reasoning that makes it possible for a human agent to reach this kind of conclusion.

The problems described above are examples of the coincidence problem which arises in a context described by an extension of Koomen's parlance as "multiple recurrence". In this paper, we address a logical formalization of multiple recurrence, and present a formalization of the solution to the problem of eventuality coincidence which arises when required to carry out planning or scheduling within such a context.

We present a reified first-order logical theory[6] that makes temporal reasoning about regular multiple recurrence possible. It is reified because the eventualities we quantify and reason about may have a hidden propositional structure from which we abstract away. Each of the recurrences in the double recurrence is regular in the sense that each of the eventualities involved has a fixed duration.  In order to keep within the confines of first order logic, we allow our eventualities to be essentially a sequence of component eventualities.  For example, we may treat "day of the week" as an eventuality comprising of each day of the week: Monday through Sunday as components. The number of components that make up an eventuality is called its length. As such we treat simple eventualities as eventualities of length 1.

Section 2 introduces the extension of Allen's interval logic used in this paper as well as the basic assumptions made about the eventualities discussed in this paper. In this paper we treat interval relations as objects in order to be able to express such second order notions as *two pairs of time interval pairs have similar qualitative relations*. Section 3 presents our formalization of the concept of multiple recurrence, which is an extension of Koomen's theory of recurrence.

We introduce the notions of periods and cycles for an eventuality's recurrence as well as the multiple recurrence of a pair of eventualities. A period is a minimal interval over which a recurring eventuality occurs once. On the other hand a cycle of the multiple recurrence of two given eventuality sequences is a minimal interval within the interval of recurrence which is started by the incidences of the two given eventuality sequences and is ended by the incidence of the two given eventuality sequences. Then we present a solution to the coincidence problem in the context of regular multiple recurrence, based on temporal projection over a cycle.  Finally, a discussion about the usefulness of the theory in the context of irregular multiple recurrence is presented.

**2 Preliminaries**

2.1 Language, Notation and Terminologies

We use a logical language to formalize the concepts described in this paper. Our logical theory is a standard first-order logic with equality. Our domains include Eventualities, Time Intervals, Time maps and Interval relations, which will include all of Allen's relations and others introduced later in this paper and the domains of ordinals and natural numbers. Our notion of a time interval is a definite length of time, without gaps (i.e. convex) and with both starting and ending instants [2, 3]. It connotes the idea of an interval of time. In some literature a time interval is regarded as a pair of time points [5]. By the notion of an interval we refer to a convex interval. As such the union of two disconnected intervals is not an interval. However, the convex union of two disconnected intervals is an interval covering the two intervals as well as its gap. A *time map* is a sequence of contiguous time intervals. Each individual time interval in a time map can be accessed by applying a variety of functions introduced later.

An *eventuality* is either a state, an event, action or a process (simple eventuality) described as a proposition or a sequence of other eventualities (complex eventuality). An example of a simple eventuality is 'it is Monday' or 'green light is on'. An example of complex eventuality is 'Day of the week' which is a sequence of week days i.e. 'Monday', 'Tuesday' etc. Another example is 'Traffic light phases' which is a sequence of Red, Amber, and Green states of the traffic light. However we do not distinguish formally between a simple eventuality and a complex one. Each eventuality has an assigned length, which designates the number of component temporal propositions. A simple eventuality has a length 1. There is an ordering among these components. Intuitively, a time- map should be understood as a sequence of contiguous time intervals, so that it is possible identify the order of the intervals in the time map. That is done in this paper by using indices, so that tm[k] represents the $k^{th}$ interval in the time- map tm. The notation tm[k] should be a syntactic denotation for a function taking a time-map and the integer k, and returning a time interval. The notion of time map used here is similar to the notion of temporal aggregates used by Pan and Hobbs [13, 14]. The only difference is that in our time map a time interval "meets" its successor in the list.

We will use the standard logical operators: conjunction $\wedge$, disjunction $\vee$, equivalence $\Leftrightarrow$, implication $\Rightarrow$ and (unary) negation $\neg$. The scope of a quantified variable is the rest of the sentence. Whenever necessary, we delimit the scope of a quantified variable by the explicit use of parentheses. The precedence of the logical operators is as assumed for first order predicate logic, i.e. precedence of $\neg$ is higher than those of $\wedge$ and $\vee$ which is higher than those of $\Leftrightarrow$ and $\Rightarrow$.

The predicates used in our theory include the standard predicates for associating eventualities with time intervals otherwise known as truth predicates. We use the predicates here as used by Koomen, i.e. *TT* (Truth), *MT* (maximal truth) and *RT* (recurrently true). We define the partition relationship, *Part* between an eventuality and a sequence of eventualities. We shall introduce the notion of partitions later. We define interval relation predicate *Intrel*, as a relation between an interval, a temporal relation and another interval. This enables us to make interval relations terms, so that we can reason about the similarity of interval relations between two pairs of intervals. The multiple recurrence predicate MRT defines the double recurrence of two eventualities over some interval. The signatures for the predicates are given below:

TT, MT, RT: Eventuality $\times$ Interval $\rightarrow$ Boolean
Intrel: Interval $\times$ Relation $\times$ Interval $\rightarrow$ Boolean
MRT: Eventuality $\times$ Eventuality $\times$ Interval $\rightarrow$ Boolean
Aux: Interval $\times$ Interval $\times$ Interval $\rightarrow$ Boolean

Of the truth predicates described above, truth (TT) is the most primitive, capturing the notion of an eventuality being true over an interval. The other truth predicates will be defined in terms of TT subsequently. Aux denotes the auxiliary interval relation. An auxiliary interval of a pair is a maximal subinterval of the convex union of the pair, other than those intervals themselves, which are not part of any interval they may have in common. For example a pair of overlapping intervals i and j has two auxiliary intervals: an interval k starting i and meeting j and another interval m met by i and finishing j. Again for a pair of disconnected intervals i and j such that i is before j, there is only one auxiliary interval and that is the interval met by i which also meets j.

In the rest of the paper, we use the letters x, y, w and z without subscripts to represent eventualities. When x, y, w, and z are made bold, they represent lists of eventualities of finite length. The symbols i, j, k and l with or without suffixes, are used for temporal intervals, while we use p, q, r, s, t, u as variable ordinals for identifying components of an eventuality. We write $w_p$ as a shorthand for the application of the component function (just ahead) to obtain the $p+1^{th}$ component of the eventualities w. Time maps are denoted by tm with or without subscripts.

The list of functions used here and their description is given below.

> dur: Interval $\rightarrow$ *N* (This function returns a natural number signifying the duration of given interval). Ontologically speaking as in [11], the notion of a duration is an amount of time measured in an appropriate unit. Our theory does not allow durations of intervals to be real numbers in the most general sense for reasons that would become clearer later.
>
> cover: Interval × Interval $\rightarrow$ Interval
> The cover function (defined in section 2.2) of two meeting intervals is the minimal interval that covers the two intervals. The function is a partial function defined only for any pair of meeting intervals which returns the minimal interval covering the two intervals.
>
> cover*: Time-map $\rightarrow$ Interval
> The cover* function (defined in section 2.2) of a time map is a single interval covering the entire time map.
>
> Π : Eventuality $\rightarrow$ Interval-set
>
> Ω : Eventuality × Eventuality $\rightarrow$ Interval-set
>
> + : Eventuality × Eventuality $\rightarrow$ Eventuality
> The infix function + is a partial function that returns an eventuality which evaluates true over any common subinterval of the incidence of the two input eventualities. The concept is somehow more general than the notion of the incidence of two eventualities over some interval. If two eventualities are downward hereditary and their incidences share a common subinterval then they are both true over that interval. The same cannot be said for two eventualities of one of any of the other classes. However x + y is true whether they are downward hereditary or not.
>
> η: Eventuality × Interval $\rightarrow$ Time-map
> The function η returns for an eventuality x, the contiguous (or meeting) sequence of its periods within the interval, if x recurs over that interval. It returns an empty list if x is not recurrently true over the given interval.
>
> Index: Time-map × *N* $\rightarrow$ Interval

The index function returns the $p^{th}$ interval of a time map, where p is the given ordinal. We will write tm[p] as a short hand for index(tm, p). Index is a partial function which is defined only when there is a $p^{th}$ interval within the time-map.

dim: Time-map $\rightarrow N$
This function returns for a given time-map, the maximum ordinal for which index will return an interval. In other words dim returns the dimension of the Time-map.

$\phi$ : Eventuality $\times$ Interval $\rightarrow$ Interval
The function returns the unique sub interval of the given interval over which the eventuality is maximally true i.e. its interval of maximal incidence within the given interval. If there is more than one such sub intervals, $\phi$ is undefined.

len: Eventuality $\rightarrow N$
The function returns the number of components that make up the eventuality. Intuitively we think of the components of an eventuality being in a sequence. A single process is regarded as a single eventuality. Thus the length of a single process is 1.

head: Time-map $\rightarrow$ Interval
The function returns the first interval in the time map.

tail: Time-map $\rightarrow$ Time-map
The function returns the list excluding the first element. It returns the value 'nil' if the time-map consist of only one interval.

comp: Eventuality $\times N \rightarrow$ Eventuality
This partial function returns the component of the given eventuality, whose position is indicated by the ordinal. As we have indicated earlier, we will write $w_p$ as shorthand for comp(w, p).

In the rest of section 2, we present an extension of Allen's interval calculus used in this paper, as well as present the basic assumptions about the kinds of eventualities we deal with in this paper.

### 2. 2 Interval Logic

Allen [2,3] formulated the interval calculus, as a means of representing time, along with a computationally efficient reasoning algorithm based on constraint propagation. Allen had argued that time points proposed earlier by McDermott [12], was inadequate.

Allen defined a linear logic of time based on the concept of intervals. In this logic, time intervals are treated as individuals. A number of qualitative binary relations are defined on intervals. The basic ones are:

| | |
|---|---|
| Meets | Finishes |
| After | Starts |
| Contains | Overlaps |

Equals[1]

The names assigned to these relations fit the relationship the reader is likely to assign to them intuitively. For example, an interval i *Meets* another interval j if i starts before j and ends just when j is starting so that they have no interval between them, and no subinterval in common. For this relation, Galton uses the term 'abut' in [8]. However the relation *After* holding between the same pair of intervals implies the existence of an interval spanning the end of i and the beginning of j.

An interval i *Contains* j if and only if the interval j starts after i starts and finishes s before i finishes. The *Equals* relation expresses the fact that i and j are the same. For the rest of the paper, we will use the equality predicate implicit in first order predicate calculus, i.e. =, instead of the Equals relation. An interval j *Finishes* i if i starts before j starts and ends with it, so that there is an interval between the start of j and the start of i. Finally, i *Overlaps* j if and only if i starts before j starts, but ends after j starts and before j ends.

Allen also built a transitivity table, which allows interval relation to be computed on a transitive basis, e.g.

> If i is *before* j and j is *before* k, then i is *before* k.
> If i is during j and j is after k, then i is after k

The transitivity relations however are not always definite. For example, if some interval j is before k and k is during l, then the relationship between j and l is one of {overlaps, *meets*, *during* and *starts*}. Again a situation in which j is before k and k is after l, suggests nothing about the relationship between j and l.

Based on these interval relations, we define (like Koomen) two additional temporal relations between time intervals: Disjoint and Within.

**Definition 2.2.1** (Disjoint relation)
*The time intervals j and k are disjoint if and only if either j is after k, or k is after j or j meets k or k meets j.*
    $\forall$j, k. Intrel(j disjoint k) $\Leftrightarrow$ Intrel(j, after, k) $\vee$ Intrel(k, after, j) $\vee$
    Intrel(k meets j) $\vee$ Intrel(j meets, k)

As usual we will write j disjoint k or k disjoint j. Note that disjoint is a symmetric relation.

**Definition 2.2.2** (Within relation)
*The interval j is within k if and only if either j starts k or j ends k or j is during k.*
    $\forall$j, k. Intrel(j within k) $\Leftrightarrow$ Intrel(j, starts, k) $\vee$ Intrel(j, finishes, k) $\vee$
    Intrel(k, contains, j)

Some times we find it necessary to refer to the subinterval relation. This relation will be a disjunction of the within relation and equality.

**Definition 2.2.3** (The Subinterval relation)

---

[1] We do not differentiate between the Allen's Equals relation and the equality predicate (=) in first order predicate calculus.

*An interval is a subinterval of another, if it is either within that interval or equal to it.*

$$\forall j, k.\ Intrel(j, sub, k) \Leftrightarrow Intrel(j, within, k) \vee k = j$$

Allen's interval relations implicitly assume the existence of some other interval apart from the related pair, whose existence is an aspect of the definition of the given relation. For example, the fact that interval j is before interval k, indicates the existence of some interval meeting k and met by j. Similarly, the fact that j starts k implies the existence of some interval met by j and ending k. If j overlaps k however, there are two auxiliary intervals: One starting j and meeting k, and another ending k and met by j. It is the existence of such interval that enabled the definition of all of Allen's interval relations in terms of the relation *meets*.

It is also important to note that auxiliary intervals defined for any pair of non-disjoint intervals, exclude any interval shared by both intervals. Those intervals are defined as common intervals as Definition 2.2.5. Thus, here we define formally the concept of an auxiliary interval for two non-disjoint intervals.

**Definition 2.2.4a** (Auxiliary Intervals)
*Given any two distinct non- disjoint intervals, an auxiliary interval is any interval, which starts or ends one of the two intervals, and meets or is met by the other.*

$$\forall j, k, m\ .\ \neg Intrel(k, disjoint, j) \wedge Aux(k, j, m) \Leftrightarrow \neg (k = j) \wedge$$
$$(Intrel(m, starts, j) \vee Intrel(m, finishes, j) \vee Intrel(j\ starts\ m)$$
$$\vee Intrel(j\ finishess\ m)\ ) \wedge$$
$$(Intrel(m, meets, k) \vee Intrel(k, meets, m))$$

**Definition 2.2.4b**
*Given any two disjoint non- meeting intervals an auxiliary interval meets one and is met by the other.*

$$\forall j, k, m\ .\ \ Intrel\ (k, disjoint, j) \wedge Aux(k, j, m) \Leftrightarrow$$
$$(\ Intrel(k\ meets\ m) \wedge Intrel(m\ meets\ j)) \vee (\ Intrel(j, meets, m) \wedge Intrel(m\ meets\ k)\ )$$

From our definition it is intuitively clear that two equal or meeting intervals have no auxiliary intervals. As such

$$\forall j, k.\ \neg \exists m.\ (Intrel(j, meets\ k) \vee j = k) \wedge Aux(k, j, m)$$

**Definition 2.2.5** (Common Intervals)
*The function common returns for any given non-disjoint interval pair their maximal common sub interval.*

$$\forall j, k\ m.\ \neg Intrel(j\ disjoint\ k) \Rightarrow$$
$$common(k, j) = m \Leftrightarrow\ Intrel(m, sub, j) \wedge Intrel(m, sub, k)$$
$$\wedge (\forall m_1.\ Intrel(m, within, m_1) \Rightarrow \neg Intrel(m_1, sub, j) \vee \neg Intrel(m_1, sub, k))$$

Now we define the idea of a time interval that covers two meeting intervals. This concept is referred to in this paper as in [10], as the cover of the two intervals.

**Definition 2.2.6** (The cover of two intervals)
*The cover of two meeting intervals is the unique interval that is started by the earlier interval and finished by the latter interval.*

$$\forall j, k, m.\ cover(j, k) = m \Leftrightarrow Intrel(j, meets, k) \land Intrel(j, starts, m) \land Intrel(k\ finishes\ m)$$

Next we formalize a basic property of time maps; namely that each interval within a time map meet the next interval within the time map. In other words every time map is a sequence of contiguous time intervals.

**Axiom 2.2.7** (Basic property of time maps)
*Each interval in a time map meets with the next interval in the map.*

$$\forall\ tm, p.\ 1 \leq p < dim(tm) \Rightarrow Intrel(tm[p]\ meets\ tm[p+1])$$

Finally, we extend the concept of cover to a time map by defining the unary function cover* for time-maps.

**Definition 2.2.8** (cover * of a time map)
*The partial function cover\* (defined for a time map) returns a single interval which covers the entire time map.*

$$\forall\ tm, m.\ cover^*(tm) = m \Leftrightarrow (dim(tm) = 1 \land m = tm[1]) \lor$$
$$(dim(tm) > 1 \land Intrel(tm[1]\ starts\ m) \land Intrel(tm[dim(tm)]\ finishes\ m))$$

This definition clearly spells out what the temporal relations between individual eventualities in the time-map tm[p] and its cover, the interval m will be.

2.3 Basic Assumptions about Eventualities

There are a number of basic assumptions about the eventualities in this paper. One of these assumptions is required in the light of existing ontological studies of the domain of eventualities. This assumption is: *All eventualities said to recur in this paper are assumed to be **contiguously repeatable***. By that we mean that two different incidences of any particular eventuality can happen over two meeting intervals. From the existing literature on the classifications of eventualities, not all eventualities have this property. For example Shoham [16]'s class of concatenable eventualities cannot be repeated. This is because by their definition, if that eventuality is ever true over two meeting intervals, then that eventuality itself (not its repetition) is true over the super interval containing the two intervals. Galton's states of position[9] are thus similar to the class of concatenable eventualities. As such they are not contiguously repeatable. Galton's states of motion on the other hand are contiguously repeatable. It seems therefore the case that the only eventualities that are contiguously repeatable are those that cannot hold true at isolated instants. Thus if there is a maximal interval over which one such eventuality holds, it cannot be incident at the interval limits. Otherwise it will not be contiguously repeatable, making it impossible to distinguish between one incidence and another incidence over a meeting interval, thus making it concatenable. A state of position such as 'x is at location l' is concatenable while a state of motion such as 'x is not at location l' is contiguously repeatable.

Another class of eventualities worth investigating for contiguous repeatability is Allen's actions [3, 4]. A particular class of actions that are repeatable are actions that are sequences of other actions which by their very nature are not concatenable and are thus contiguously repeatable. For example, the action of "Hitting a nail on its head with a hammer" consists of two actions:

1. Raising the hammer
2. Moving the hammer to the nail's head with some momentum.

This composite action is contiguously repeatable. Similarly while a heating action is concatenable, a sequence of heating and cooling actions is contiguously repeatable.

Our second assumption is that *every incidence of any specific eventuality has a fixed duration* as opposed to a variable or a fuzzy duration. Calendar eventualities such as days of the week (e.g. Monday..) are of fixed duration. So are timed or synchronized actions or processes. However some actions and processes have an upper and a lower limit on their duration. For example: *a forging process takes between 3 and 4 hours*. These are said to have a fuzzy duration. An eventuality with variable duration is neither fixed nor fuzzy. It is not fixed and has neither upper nor lower limits. For example we have no idea how long *a nailing process* is. It depends on the hardness of the wall.

In what follows section 3 we introduce Truth (TT), Maximal truth (MT) and Recurrence.

## 3. Maximal Truths, Sequences and Recurrences

In the literature, one of the reasons for wishing to reason about time is to be able to reason about the incidences of states, events, processes, actions, *etc.* over time. The propositions representing these eventualities are not necessarily timelessly true. Thus we refer to them as eventualities. Different authors have used different predicates for expressing the incidence of eventualities on time units. The predicate *Holds*, has been used for properties and states, while *Occurs* have been used for events [2].

A generalized predicate we will use for all eventualities irrespective of their specific nature is *TT* intended to mean true. Thus one can take TT(x, j) to mean that x is true over the interval j. This kind of truth notion should be contrasted with *true within* which has also appeared in the literature expressed as the predicates *Holds-in* and *Occurs-in* for states and events respectively [9].

Koomen also defined the concept of maximal truth using the predicate *MT*. Maximal truth was introduced as a means of preventing infinite truths of eventualities, over time intervals. By Koomen's description, to say, x is maximally true over an interval j, means x is true throughout the interval j and it is not true over any interval k, which intersects or is a super interval of j. However our TT predicate is intended to mean true over as opposed to true throughout which is only appropriate for a few classes of eventualities such as Shoham's class of downward hereditary whose truth over an interval implies truth over all its subintervals. Thus we write MT(x, j) to refer to the fact that x is maximally true over an interval j.

However since our concept of eventuality is more intrinsically complex than Koomen's, we need to define the incidence of an eventuality over an interval in terms of the incidences of its components over parts of the interval. Consequently we will describe the concept of maximal truth in two stages; first defining maximal truth of a proposition describing an eventuality with a single component, and then defining maximal truth of a proposition describing eventualities with many components.

**Axiom 3.1a** (Maximal truth of eventualities of length 1)
*A simple eventuality x is maximally true over an interval j if x is true over j and x is not true over any interval which is a super interval of j, or overlaps or is overlapped by j.*

$\forall$ x, k . length(x) =1 $\Rightarrow$
(MT(x, k) $\Leftrightarrow$ TT(x, k) $\land$ ($\forall$j. Intrel(j, overlaps, k) $\lor$ Intrel(k, Overlaps j) $\lor$
Intrel(k, within, j) $\Rightarrow$ $\neg$ TT(x, j)))

**Axiom 3.1b** (The truth of eventualities of length greater than 1)
*If an eventuality x of length greater than 1 is true over an interval j, then its first component $x_0$ is maximally true over some starting subinterval of j, and provided the length is at least p, the p-1$^{th}$ component is maximally true over a subinterval of j which meets another subinterval of j over which the $k^{th}$ component is maximally true.*

$\forall$x,k. len(x) > 1 $\Rightarrow$
( TT(x, k) $\Leftrightarrow$ $\exists$ $k_1$. MT($x_0$, $k_1$) $\land$ Intrel($k_1$ starts k) $\land$
($\forall$p. 0 < p $\leq$ length(x) -1 $\Rightarrow$ $\exists k_2, k_3$. Intrel($k_2$, within, k) $\land$ Intrel($k_3$, within, k) $\land$
MT($x_{p-1}$, $k_2$) $\land$ MT($x_p$, $k_3$) $\land$ Intrel($k_2$ meets $k_3$) )

**Axiom 3.1c** (The maximal truth of eventualities of length greater than 1)
If an eventuality is of length greater than 1, then it is maximally true over an interval when it is true over the same interval.

$\forall$x, k. len(x) > 1 $\Rightarrow$
MT(x, k) $\Leftrightarrow$ TT(x, k)

We are treating eventuality sequences as a solid entity and as such they are only true over intervals that they are maximally true over.

The next theorem establishes the relationship between truth and maximal truth for all eventualities. It follows from definitions 3.1a and 3.1c.

**Theorem 3.2**
*If an eventuality x is maximally true over an interval j, it is also true over the same interval.*
$\forall$x. MT(x, j) $\Rightarrow$ TT(x, j)

A basic domain assumption is the fact that a proposition defining an eventuality cannot be true forever. The next Axiom formalizes this domain assumption.

**Axiom 3.3** (Nothing is true forever)
*If an eventuality x is true over an interval j, then j is a part of or equal to some interval k over which x is maximally true.*
$\forall$x, j. TT(x, j) $\Rightarrow$ $\exists$k. MT(x, k) $\land$ (j = k $\lor$ Intrel(j within k))

The next axiom introduces a slight restriction needed to achieve the representation of regular recurrence. While this may slightly limit the applicability of the domain, as we have seen before in the examples in section 1, there are domains where this restriction does not limit.

**Axiom 3.4** (Fixed duration eventualities)
*If an eventuality x is maximally true over two different intervals j and k, then the durations of j and k are equal.*

$$\forall\, x, j, k.\ MT(x, j) \wedge MT(x, k) \Rightarrow dur(j) = dur(k)$$

Such an assumption will be valid when we are dealing with eventualities with fixed duration of maximal incidence, as opposed to variable durations. All of the results of this paper are based on this assumption. As such, the results discussed here in section 3 and 4 will be invalid in a domain where there is some fuzziness about how long a state or event may take. An example of fuzziness is when an event takes between 2 and 3 hours to complete. Interestingly, there are applications for which we do not have to contemplate the problem of fuzziness. However we return to the problem of fuzzy durations later.

We now define the truth of an eventuality x+y which is defined as the coincidence of x and y. As introduced earlier + when applied to a pair of eventualities returns an eventuality that corresponds to the incidences of the two eventualities sharing a common subinterval. However + can only be formally defined in terms of the truth of the product of its application. That is presented here:

**Definition 3.5**  (+ operators for eventualities)
a. *We say that the eventuality x + y is maximally true over an interval if and only if that interval is a common subinterval of two non-disjoint intervals such that x is maximally true over one of the intervals and y is maximally true over the other.*

$$\forall x, y, k.\ MT(x + y, k) \Leftrightarrow \exists j, i.\ MT(x, j) \wedge MT(y, i) \wedge \neg Intrel(i, disjoint\ j)$$
$$\wedge\ k = common(j, i)$$

b. *Eventuality x + y is true over an interval k if and only if k is a subinterval of some other interval over which x + y is maximally true.*

$$\forall x, y, k.\ TT(x + y, k) \Leftrightarrow \exists\, j.\ MT(x + y, j) \wedge Intrel(k, subs, j)$$

The following Theorem contains some properties of the + operator. All of the properties follow from Definition 3.5.

**Theorem 3.6** (Properties of +)
a. *The operator + is commutative.*
$$\forall x, y, k.\ MT(x+y, k) \Leftrightarrow MT(y+x, k)$$

b. *For all x and y, x + y is downward hereditary irrespective of the Shoham type of x and y..*
$$\forall x, y, k.\ MT(x+y, k) \Rightarrow \forall j.\ Intrel(j\ within\ k) \Rightarrow TT(x+y, j)$$

c. *If x+y+z is maximally true over any interval then x+y, y+z and x+z are all true over that interval.*
$$\forall x, y, z, j.\ MT(x+y+z, j) \Rightarrow MT(x+y, j) \wedge MT(x+z, j) \wedge MT(z+y, k)$$

It is appropriate at this point to introduce the ϕ (phi) function which returns a unique interval of incidence, within an interval if indeed there is such an interval, and is not defined if there is no such interval or there is more than one such.

**Definition 3.7** (The phi (ϕ) function)
*Given an eventuality x and interval k, the ϕ function returns the only (possibly improper) subinterval of k, over which x is maximally true. It is undefined either if there is no such interval or there are more than one such interval.*

$$\forall x, k, j.\ \phi(x, k) = j \Leftrightarrow \exists! i.\ MT(x, i) \wedge Intrel(i, sub, k) \wedge j = i$$

Maximal truth also forms the basis for defining recurrence. Although Koomen offered no formal definition of recurrence but had an axiom that states that: if x recurs over an interval then *either* it is maximally true over the interval *or* it is maximally true over some starting subinterval of the interval, and it recurs over the remaining subinterval of the given interval. However we strengthen that axiom and offer a definition for recurrence thus:

**Definition 3.8**(Recurrence)
*An eventuality is said to recur over an interval if and only if either it is maximally true over the interval or it is maximally true over some starting subinterval of the interval, and it recurs over the remaining subinterval of the given interval.*

$$\forall x, k.\ RT(x, k) \Leftrightarrow MT(x, k) \vee$$
$$(\exists j, i.\ Starts(j, k) \wedge Ends(i, k) \wedge Meets(j, i) \wedge MT(x, j) \wedge RT(x, i))$$

The effect of this strengthening is to make maximal truth a special case of recurrence.

Now we introduce the η function into our formalism. We use the function to refer to the list of intervals over which x is maximally true, within an interval over which x recurs.

**Axiom 3.9**(The eta η function)
*Given that x recurs over an interval j, then, the function η returns a time map of the interval whose cover* is the original interval, and x is maximally true over each time interval returned by the index function.*

$$\forall x, j.\ RT(x, j) \Rightarrow$$
$$\exists tm.\ \eta(x, j) = tm \wedge j = cover^*(tm) \wedge (\forall p.\ 1 \leq p \leq dim(tm) \Rightarrow MT(x, tm[p]))$$

The eta function η is a partial function that takes an eventuality and an interval over which the eventuality is recurrently true, and returns a time map which contains time intervals over which the eventuality is true throughout. To illustrate the usefulness of eta function in our formalization of recurrence, we can express the fact that if x is recurrently true over k, then any subinterval j of k over which x is maximally true, is a member of the list returned by η(x, k) by writing:

$$\forall x, k.\ RT(x, k) \Rightarrow (\forall j.\ MT(x, j) \wedge Intrel(j\ sub\ k) \Rightarrow \exists\ n1.\ \eta(x, k)[n1] = j)$$

This statement above is valid and it is generalized for multiple recurrence later in Theorem 3.21b.

**Definition 3.10**(Multiple Recurrence)
*We say that x and y doubly recur over an interval k if and only if both x and y each recur over k*
$$\forall x, y, k.\ MRT(x, y, k) \Leftrightarrow RT(x, k) \wedge RT(y, k)$$

So for example, to say x recurs over an interval j in Koomen's theory, one writes as a logical formula RT(x, j). However, there is often the need to represent the recurrence of two different eventualities over the same interval. For example, to state that eventualities x and z recur over the interval j we will write the formula: RT(x, j) and. RT(z, j). We will refer to this as a double recurrence of x and z over j.

The following is basically true of components of an eventuality because any two of them cannot be maximally incident on two non-disjoint intervals.

**Axiom 3.11(Mutual exclusion)**
*Any two different components of an eventuality are mutually exclusive in time, i.e. any pair of intervals over which they respectively true, must be disjoint.*
$$\forall\ x, p, q, j, k.\ p \neq q \wedge TT(x_p, j) \wedge TT(x_q, k) \Rightarrow Intrel(k, disjoint, j)$$

We proceed to define some properties of recurrence and multiple recurrence, and some basic truths about them.

3.1 Properties of Recurrence and Multiple Recurrence

This section presents basic inferences about the properties of recurrence and double recurrence. The notions of periods of an eventuality and the cycles of a multiple recurrence are introduced. Their properties are also proved.

**Definition 3.13** (Period of Eventuality)
*A period of an eventuality is an interval over which it is maximally true. We assume that the function $\Pi$ returns the set of all such intervals for any given eventuality.*

$$\forall x, \pi, k.\ \pi \in \Pi(x) \Leftrightarrow MT(x, \pi)$$

The following theorem follows from the definitions of recurrence (3.8) and multiple recurrence (3.9). The proof is trivial and is thus omitted.

**Theorem 3.14**
*If there is a double recurrence of x and y over an interval k, then Either there exists two starting sub intervals of k such that x is maximally true over one and y is maximally true over the other and two ending subintervals of k such that x is maximally true over one and y is maximally true over the other Or either x or y are maximally true over k.*

$\forall k, x, y.\ MRT(x, y, k) \Rightarrow$
$(\exists\ j_1, j_2, j_3, j_4.\ Intrel(j_1, starts\ k) \wedge$
$\quad (\ Intrel(j_2, starts, k) \wedge MT(x, j_1) \wedge MT(y, j_2) \wedge$
$\quad\quad MT(x, j_3) \wedge MT(y, j_4) \wedge Intrel(j_3\ ends\ k) \wedge Intrel(j_4, ends\ k)) \vee$
$\quad (\ MT(x, k) \vee MT(y, k)\ )$

Now we define the cycle of a double recurrence.

**Definition 3.15** (Cycle of a double recurrence $\Omega$)
*A cycle $\omega$ of a double recurrence of two eventualities x and y, is a minimal interval over which x and y are doubly recurrent.*

$$\forall x, y, \omega.\ \omega \in \Omega(x, y) \Leftrightarrow$$

$$MRT(x, y, \omega) \wedge$$

$$\forall j.\ Intrel(j, within, \omega) \Rightarrow j \notin \Omega(x, y)$$

The following theorem follows directly from the definitions of multiple recurrence(3.10) and cycles of multiple recurrence (3.15) as well as Theorem 3.14.

**Theorem 3.16**
*If two eventualities x and y are recurrently true over an interval k, and there is a cycle of the multiple recurrence $\omega$ which is a subinterval of k then either $\omega = k$ or there exists an interval $k_1$, such that $k = cover(\omega, k_1)$.*

$$\forall x, y, k, \omega.\ MRT(x, y, k) \wedge \omega \in \Omega(x, y) \wedge Intrel\ (\omega\ sub\ k) \Leftrightarrow$$
$$(\omega = k \vee \exists k_1.\ cover(\omega, k_1) = k \wedge MRT(x, y, k_1))$$

The next axiom presents a necessary condition for an interval to share a common subinterval with an interval over which some eventuality recurs.

**Theorem 3.17**
*If some interval k which shares a common subinterval with an interval over which an eventuality x recurs, then there exists a period of x with which y shares a common subinterval.*

$$\forall x, j, k.\ RT(x, j) \wedge \neg Intrel(j\ disjoint, k) \Rightarrow$$
$$\exists n_1. \neg Intrel(\eta(x, j)[n_1], disjoint, k)$$

*Proof*
This theorem can be proved by contradiction thus. Assume the left hand side of the implication is true while the right hand side is not. In that case $\neg \exists n_1. \neg Intrel(\eta(x, j)[n_1], disjoint, k)$. Hence it holds true that $\forall n_1. Intrel(\eta(x, j)[n_1], disjoint, k)$. If that is the case then, j and k are disjoint because by axiom 3.9, $j = cover^*(\eta(x, j)[n_1])$. This last conclusion contradicts the right hand side of the implication and the proof is concluded•.

The following Theorem 3.18 is a direct consequences of our basic assumption i.e. Axiom 3.4 and the Definition 3.13 of periods.

**Theorem 3.18**
*All periods of the same eventualities have the same durations.*
$$\forall x, \pi_1, \pi_2.\ \pi_1, \pi_2 \in \Pi(x) \Rightarrow dur(\pi_1) = dur(\pi_2)$$

The following Theorem 3.19 is a direct consequence of Axiom 3.14 and the Definition 3.8 of recurrence.

**Theorem 3.19**
*If an eventuality x recurs over an interval k, then the duration of k is a multiple of the duration of maximal incidence of j.*

$$\forall x, k.\ RT(x, k) \Rightarrow \forall \pi_1.\ \pi_1 \in \Pi(x) \Rightarrow \exists r_1.\ dur(k) = r_1 * (dur(\pi_1))$$

The following theorem is a direct consequence of Axiom 3.4 and the Definition 3.15 of a cycle.

**Theorem 3.20**
*All cycles of double recurrence for the same pairs of eventualities are of the same duration.*

$$\forall x, y, \omega_1, \omega_2.\ \omega_1, \omega_2 \in \Omega(x, y) \Rightarrow dur(\omega_1) = dur(\omega_2)$$

In all cases Theorems 3;18-3.20, the proofs are trivial and left to the reader to figure out.

Both theorems 3.21 (a and b) follow from Axiom 3.9 introducing η and Definition 3.15 defining cycles of multiple recurrence.

**Theorem 3.21**
*a. If ω is a cycle of the double recurrence of eventualities x and y, then ω is a sequential composition of the periods of x that are within it, as it is the sequential composition through cover\* of the periods of each of x and y that are within it.*

$$\forall \omega, x, y.\ \omega \in \Omega(x, y) \Rightarrow \omega = cover^*(\eta(x,\omega)) \wedge \omega = cover^*(\eta(y,\omega))$$

*b. For any cycle ω of the multiple recurrence of sequences x and y, if we know that $x_p$ and $y_q$ are maximally true over some intervals $k_1$ and $k_2$ respectively within ω, then $k_1$ is the interval of maximal incidence of $x_p$ within some period of x within ω and $k_2$ is the interval of maximal incidence of $y_q$ within some period of y within ω.*

$$\forall x,\ p,\ \omega,\ y,\ k_1,\ k_2.\ \omega \in \Omega(x, y) \wedge MT(x_p, k_1) \wedge MT(y_q, k_2) \wedge Intrel(k_1, within, \omega) \Rightarrow$$
$$\exists n_1, n_2.\ k_1 = \phi(x_p, \eta(x, \omega)[n_1]) \wedge k_2 = \phi(y_q, \eta(y, \omega)[n_2])$$

Theorem 3.21a follows from the Definition 3.15 of ω and 2.2.8 of cover\*. Theorem 3.21b follows from Definitions 3.10 of MRT and 3.15 of ω as well as Axiom 3.9.

The final theorem in this section follows from the fact of axiom 3.4 about fixed durations of each incidence of eventualities and some arithmetic reasoning on the definition of a cycle (3.15).

**Theorem 3.22**
*The duration of a cycle of double recurrence equals the least common multiple of the periods of the two recurrences.*

$$\forall x, y, \omega, j, k, n_1, n_2.$$
$$\omega \in \Omega(x, y) \wedge (MT(x, j) \Rightarrow dur(k) = n_1) \wedge (MT(y, k) \Rightarrow dur(k) = n_2) \Rightarrow$$
$$dur(\omega) = lcm(n_1, n_2)$$

The proofs of Theorems 3.21 and 3.22 are also left out.

3.2 Formal statement of the "coincidence" problem and a solution

We are now better equipped to define the coincidence problem more formally. The problem statement is presented thus:

*Given two sequences of eventualities seq($x_0,..x_{s-1}$) and seq($y_0..y_{t-1}$) recurring over some interval j, which is longer than a cycle. For some p and q within the bounds [0..s-1] and [0..t-1] respectively, can we find some subinterval of j over which both $x_p$ and $y_q$ are true ?*

Using the logical language we have developed our problem is to infer the truth or falsity of the following assertion

> Given particular sequences x and y and interval k such that MRT(x, y, k) and ordinals p and q within the limits of the ordinals in the sequences
>
> Is it the case that: $\exists j.\ Intrel(j, within, k) \wedge TT(x_p + y_q, j)$ ?

The question posed above can be answered by the following theorem.

It is intuitively clear that every cycle is exactly like others. By this we mean that given an x-sequence **x** and y-sequence **y** say, and given any two respective members of the sequences $x_p$ and $y_q$, both recurring over some interval, then the intervals of maximal incidence of $x_p$ within the $r^{th}$ period of the x-sequence, and interval of maximal incidence of $y_q$ within the $s^{th}$ period of recurrence of the y-sequence, have a temporal relationship that is preserved from cycle to cycle.

We formalize this with the next theorem and corollary. The first one states that the $r^{th}$ interval of maximal incidence of the any given member of the sequence x and the $s^{th}$ interval of maximal incidence of any given member of y have an interval relation that is fixed from cycle to cycle.

**Theorem 3.23** (Each cycle is exactly like every other 1)
*The $r^{th}$ period of x within a cycle of x and y and the $s^{th}$ period of y within same have the same relationship in any cycle.*

$$\forall \omega_1, \omega_2, x, y.\ \omega_1, \omega_2 \in \Omega(x, y) \Rightarrow$$
$$\forall r, s\ rel_1\ rel_2.\ Intrel(\eta(x, \omega_1)[r]\ rel_1\ \eta(y, \omega_1)[s]) \wedge$$
$$Intrel(\eta(x, \omega_2)[r]\ rel_2\ \eta(y, \omega_2)[s]) \Rightarrow rel_1 = rel_2$$

(The Proof of Theorem 3.23 is contained in the appendix)

**Corollary 3.24** (Each cycle is exactly like every other 2)
*For the double recurrence of two sequences* x *and* y*, the relation between interval of incidence of any two members of the sequences $x_p$ and $y_q$ say, within the $r^{th}$ period of the x-sequence and the $s^{th}$ period of the y-sequence respectively, are the same in all cycles.*

$$\forall \omega_1, \omega_2, x, y.\ \omega_1, \omega_2 \in \Omega(x, y)) \Rightarrow$$
$$\forall p, q, r, s, rel_1, rel_2.\ 0 \leq p \leq length(x) -1 \wedge 0 \leq q \leq length(y) -1 \wedge$$
$$Intrel(\phi(x_p, \eta(x, \omega_1)[r])\ rel_1\ \phi(y_q, \eta(y, \omega_1)[s])) \wedge$$
$$Intrel(\phi(x_p, \eta(x, \omega_2)[r])\ rel_2\ \phi(y_q, \eta(y, \omega_2)[s])) \Rightarrow rel_1 = rel_2$$

Corollary 3.24 is also justified in a way similar to 3.23. The facts listed 2 to 4 in the proof made in respect of Theorem 3.23, are also true for both of the pairs: $\phi(x_p, \eta(x, \omega_1)[r])$ $\phi(y_q, \eta(x, \omega_1)[r])$ and $\phi(x_p, \eta(x, \omega_2)[r])$ $\phi(y_q, \eta(x, \omega_2)[r])$ as is fact 1 in respect of $\omega_1, \omega_2$.

We now conclude this section by providing a solution to the problem.

**Theorem 3.25** (A cycle is enough projection to determine coincidence)
*If eventualities x and y are recurrently true over some interval k, then $x_p + y_q$ is true over some subinterval of k if and only if $x_p + y_q$ can be shown to be true over some interval within any arbitrary cycle of the multiple recurrence within k.*

$\forall x, y, p, q, r, s.$
$MRT(x, y, k) \land 0 \leq p \leq length(\mathbf{x}) -1 \land 0 \leq q \leq length(\mathbf{y}) -1 \Rightarrow$
$\exists j, Intrel(j, within, k) \land MT(x_p + y_q, j) \Leftrightarrow$
$\exists r, s, \omega. \omega \in \Omega(\mathbf{x}, \mathbf{y}) \land Intrel(\omega, sub, k) \land$
$\quad \neg Intrel(\phi(x_p, \eta(x, \omega)[r]), disjoint, \phi(y_q, \eta(y, \omega)[s]))$

(The Proof of Theorem 3.25 is contained in the appendix)

The implication of this theorem is that to determine the existence of an interval in which both $x_p$ and $y_q$ are true, with an interval of double recurrence of the sequence x and y, all we need to do is to determine if any such interval exists within any arbitrary cycle of the double recurrence of x and y. If such an interval exists then there is an interval of coincidence of $x_p$ and $y_q$ within any interval of double recurrence of x and y that is at least as long as a cycle of the double recurrence. As such a projection algorithm for determining coincidence among two sequences x and y has a worst case running time of $O(len(x)*len(y))$.

## 4. Relaxing the Restrictions

There is a need to justify the use of this logical apparatus for solving the foregoing problem, particularly considering the fact that the durations of the incidence of eventualities are known ahead of time. The use of a qualitative logic such as Allen's interval logic and its variants is justified by the fact that there is a need to be mindful of how to integrate the solution to this problem with existing solutions of other temporal reasoning problems. An adequate logical theory for solving many of the reasoning problems that may arise from recurrence and multiple recurrence is needed. As such, the logic described in this paper is capable of solving the same reasoning problem addressed by Koomen in addition to the problem of coincidence solved here.

The logical theory presented here can also be amended to represent and reason with multiple recurrence in a wider context, when the durations of maximal occurrence of eventualities are not fixed. Note that when durations are variable, the notion of a cycle is no longer as intuitively understood in regular periodic recurrence. If we remove Axiom 3.4 and its consequents such as Theorems 3.18-3.20 as well as Theorems 3.23 and 3.25 from the theory, we still have a theory that can represent and reason with multiple recurrence. For example where expected durations can be computed or otherwise determined for each interval of occurrence of each eventuality ahead of time during temporal projection, then the theory can predict coincidence if multiple recurrence happened over some fixed duration. For example one might be able to predict the expected duration of $\phi(x_p, \eta(x, j)[r])$, (within any interval j over which x recurs) through a probability distribution function of p and r.

Even when durations of interval of incidence for eventualities are variable and cannot be predicted, this logical theory still has the potential to be applied in monitoring with a view to averting an undesired coincidence of two eventualities from a sequence. For instance if we wish to avert the coincidence $x_p$ and $y_q$ within an interval over which the multiple recurrence of x and y happen, what we need to do is to allow time points or instants in the same way that Galton [9] has done by introducing the functions rlimit and llimit which return the right and left limits of time intervals respectively. So that the following axiom is true:

**Axiom 4.1**
*The right limit of any incidence of an element of an eventuality within an occurrence of an eventuality sequence of which it is part is the same instant as the left limit of the incidence of its successor within the eventuality sequence.*

$$\forall x, j, r, t.\ \text{rlimit}(\phi(x_{t-1}, \eta(x, j)[r])) = \text{llimit}(\phi(x_t, \eta(x, j)[r])$$

The Axiom 4.2 below allows one eventuality to disable the other, while it is holding. This axiom disables $y_q$ when $x_p$ becomes true.

**Axiom 4.2**
*If a multiple recurrence of x and y is true over an interval j and we are trying to avert the situation in which incidences of $x_p$ and $y_q$ are non-disjoint, then if a trigger is received for the beginning of an incidence of $x_p$ and $x_p$ is not currently disabled, then disable $y_q$.*

$\forall\ x, y, p, q.\ \text{MRT}(x, y, j) \wedge 1 < p \leq \text{len}(x) \wedge 1 \leq q \leq \text{len}(y) \wedge \text{Averting}(x_p, y_q) \Rightarrow$
$\exists r\ .\ \text{Now} = \text{rlimit}(\ \phi(x_{p-1}, \eta(x, j)[r])) \wedge$
$\neg(\exists t.\ \text{Intrel}(t\ \text{before Now}) \wedge \text{Disable}(\ x_p, t) \wedge (\forall t'\ t \angle t' \angle \text{Now} \Rightarrow \neg\text{Clip-Disable}(x_p, t')))$
$\Rightarrow \text{Disable}(\ y_q,\ \text{Now})$

'Now' is a 0-ary function returning current time point and $\angle$ is a linear ordering among time points known as 'before'. A similar axiom is needed to disable $x_p$ when $y_q$ becomes true.

**Axiom 4.3**
*If a multiple recurrence of x and y is true over an interval j and we are trying to avert the situation in which incidences of $x_p$ and $y_q$ are non-disjoint, then if a trigger is received for the beginning of an incidence of $y_q$ and $y_q$ is not currently disabled, then disable $x_p$.*

$\forall\ x, y, p, q.\ \text{MRT}(x, y, j) \wedge 1 \leq p \leq \text{len}(x) \wedge 1 < q \leq \text{len}(y) \wedge \text{Averting}(x_p, y_q) \Rightarrow$
$(\exists r\ .\ \text{Now} = \text{rlimit}(\ \phi(y_{q-1}, \eta(y, j)[r])) \wedge$
$\neg(\exists t.\ \text{Intrel}(t\ \text{before Now}) \wedge \text{Disable}(\ y_q, t) \wedge (\forall t'\ t \angle t' \angle \text{Now} \Rightarrow \neg\text{Clip-Disable}(y_q, t'))))$
$\Rightarrow \text{Disable}(\ x_p,\ \text{Now})$

Please note that the disabling of an eventuality is done at a point in time and persists until it is 'clipped' as it is done in many places in the AI literature e.g. [5].

In order to avoid a deadlock situation in which both eventualities disable each other we have the Axiom 4.4:

**Axiom 4.4**

*If a multiple recurrence of x and y is true over an interval j and we are trying to avert the situation in which incidences of $x_p$ and $y_q$ are non-disjoint, then if triggers are received at the same instant for the beginning of an incidence of $x_p$ and $y_q$, then an the system must halt for external intervention.*

$$\forall\ x, y, p, q.\ MRT(x, y, j) \wedge 1 < p \leq len(x) \wedge 1 < q \leq len(y) \wedge Averting(x_p, y_q) \Rightarrow$$
$$(\exists r, s\ .Now = rlimit(\phi(x_{p-1}, \eta(x, j)[r])) \wedge Now = rlimit(\phi(y_{q-1}, \eta(y, j)[s])))$$
$$\Rightarrow Halt\text{-}Check1(Now)$$

Finally, we can make the main logic described in this paper solve the problem of coincidence when the durations are rational numbers. The durations can be treated as integers by changing the standard unit by a factor of $10^{-k}$ where k is the maximum number of decimal places in any of the given durations. For example, if the duration of each occurrence of x is 12 and that of the occurrence of y is 13.5, the by altering the unit of durations by $10^{-1}$, then duration of an occurrence of x becomes 120 and that of y becomes 135. However, when either duration of the eventualities involved in the multiple recurrence is irrational, then it becomes impossible to make its duration a natural number by changing the global unit of durations. As such, the solution proposed here is only applicable when durations of intervals are rational

## 5. Summary and Conclusions

We have presented a logical formalization for reasoning about multiple recurrence, and thereby present a simple solution to the coincidence problem. The importance of the coincidence problem has been discussed. Based on the formalization of multiple recurrence, we have presented a solution to the coincidence problem. The solution is contained in Theorems 3.25.

The solution presented in this paper projects the eventualities over a cycle of multiple recurrence for x and y. If a coincidence does not occur within a cycle, then it will never occur. The running time of the projection algorithm proposed for solving the coincidence problem is quadratic.

Multiple recurrence occurs in a number of real life scenarios. As such it is worthwhile to pursue efficient algorithmic solutions to the problem of coincidence within the context of multiple recurrence.

**Acknowledgements**
The author is most grateful to God for another chance at life. Without Him, this paper would never have been written or published. He is also grateful to the referee whose insightful comments on earlier drafts results in the current form of this manuscript. He is also grateful to family, friends, colleagues and brethren whose help and support during a very trying time has proved very critical to the outcome of this effort. I am not sure that *one man has ever owed so much to each of so many people*.

# Appendix: Proofs

## *Proof of Theorem 3.23*

In order to prove theorem 3.23, we need a number of lemmas here. The concept of similar relationship between two pairs of intervals is presented. We then prove that the relationships between $\eta(x, \omega)[p]$ and $\eta(x, \omega)[q]$ is similar from cycle to cycle.

**Definitions 3.23a**
Two pairs of intervals $(x, y)$ and $(x_1, y_1)$ are said to be similar if and only if

- (i) x and x1 have the same duration and y and y1 have the same duration.
- (ii) The relationship between x and y is exactly the same as that between $x_1$ and $y_1$, i.e.

$$\forall rel, rel_1. \; Intrel(x, rel\; y) \wedge Intrel(x_1, rel\; y_1) \Rightarrow (rel = rel_1)$$

- (iii) For each k such that Aux(x, y, k), there exists some interval k1 such that Aux(x1, y1, k1) such that k and k1 have the same durations and the relations between x and k is the same as that between x1 and k1. Similarly the relation between k and y is the same between k1 and y1.

**Lemma 3.23b**
$$\forall x, x_1, y, y_1. \; Similar((x, y), (x_1, y_1)) \Rightarrow Similar((y, x), (y_1, x_1))$$

**Lemma 3.23c**
If two pairs of intervals $(x, y)$ and $(x_1, y_1)$ are similar, then any two new pairs $(w, z)$ and $(w1, z1)$ such that x meets w and y meets z and x1 meets w1 and y1 meets z1 and $dur(w) = dur(w1)$ and $dur(z) = dur(z1)$ are similar.

**Lemma 3.23d**
If two pairs of intervals $(x, y)$ and $(x1, y1)$ are similar and there are two pairs of intervals $(w, z)$ and $(w1, z1)$ such that x meets w and y meets z and x1 meets w1 and y1 meets z1 and $dur(w) = dur(w1)$ and $dur(z) = dur(z1)$., then it is the case that
$\quad Similar((x, z), (x_1, z_1))$ and $Similar((y, w), (y_1, w_1))$

**Lemma 3.23e**
If $Similar((x, y), (x_1, y_1))$ and $Similar((x_1, y_1), (x_2, y_2))$ then $Similar((x, y), (x_2, y_2))$.

$$\forall x,y,x_1,y_1. \; Similar((x, y), (x_1, y_1)) \wedge Similar((x_1, y_1), (x_2, y_2)) \Rightarrow Similar((x, y), (x_2, y_2))$$

**Lemma 3.23f**
For all h intervals $\omega_1, \omega_2$ such that: $\omega_1, \omega_2 \in \Omega(x, y)$ it is the case that:
$\quad Similar(\;(\eta(x, \omega_1)[1], \eta(y, \omega_1)[1]), (\eta(x, \omega_2)[1], \eta(y, \omega_2)[1])\;)$

Thus to put it in a very informal language, $\eta(x, \omega_1)[1]$ and $\eta(x, \omega_2)[1]$ start and end at similar points within $\omega_1$ and $\omega_2$ and indeed any cycle. Similarly for $\eta(y, \omega_1)[1]$ and $\eta(y, \omega_2)[1]$. The intervals $\eta(x, \omega_1)[1]$ and $\eta(y, \omega_1)[1]$ start at the same time. Similarly $\eta(x, \omega_2)[1]$ and $\eta(y, \omega_2)[1]$ start at the same time. Note that Therefore, $\eta(x, \omega_1)[1]$ and $\eta(y, \omega_1)[1]$ have a similar temporal relationship to $\eta(x, \omega_2)[1]$ and $\eta(y, \omega_2)[1]$.

**Lemma 3.23g**

For all integers p and q within limits and intervals $\omega_1, \omega_2$ such that: $\omega_1, \omega_2 \in \Omega(x, y)$ it is the case that:

$\text{Similar}( (\eta(x, \omega_1)[p], \eta(y, \omega_1)[q]), (\eta(x, \omega_2)[p], \eta(y, \omega_2)[q]) )$

**Proof**:

1. It is the case that $\text{Similar}( (\eta(x, \omega_1)[1], \eta(y, \omega_1)[1]), (\eta(x, \omega_2)[1], \eta(y, \omega_2)[1]) )$ from Lemma 3.23e.

2. It is also the case that $\text{Similar}( (\eta(x, \omega_1)[1], \eta(y, \omega_1)[q]), (\eta(x, \omega_2)[1], \eta(y, \omega_2)[q]) )$ for all values of q within limits.

   (This follows from step 1 above and a repeated application of Lemma 3.23c and 3.23e)

3. It is also the case that $\text{Similar}( (\eta(x, \omega_1)[2], \eta(y, \omega_1)[q]), (\eta(x, \omega_2)[2], \eta(y, \omega_2)[q]) )$ for all values of q within limits.

   (This follows from Lemma 3.23d and 3.23b and then a repeated application of 3.23 c and 3.23e.)

4. For any arbitrary k within limits, it is the case that
   $\text{Similar}( (\eta(x, \omega_1)[k], \eta(y, \omega_1)[q]), (\eta(x, \omega_2)[k], \eta(y, \omega_2)[q]) )$ for all values of q from 2 on.

   (This follows from a repeated application Lemma 3.23d and 3.23b for values of q that
   $\eta(y, \omega)[q]$ starts before $\eta(x, \omega)[k]$ in any cycle,
   And from a repeated application of Lemma 3.23c and 3.23 e for other values of q) •

**Main Proof of Theorem 3.23**

1. From Lemma 3.23g it is clear that If $\omega_1, \omega_2 \in \Omega(x, y)$ it is the case that:
   $\text{Similar}( (\eta(x, \omega_1)[r], \eta(y, \omega_1)[s]), (\eta(x, \omega_2)[r], \eta(y, \omega_2)[s]) )$ for all r and s within limits.

2. From the definition of Similarity particularly 3.23a (ii) it is the case:
   $\forall x, x_1, y, y_1. \text{Similar}((x, y) (x_1, y_1)) \Rightarrow \forall \text{rel}, \text{rel}_1. \text{Intrel}(x, \text{rel } y) \wedge \text{Intrel}(x_1, \text{rel } y_1)$
   $\Rightarrow (\text{rel} = \text{rel}_1)$

3. From 1 and 2 and modus ponens :
   $\text{Intrel}(\eta(x, \omega_1)[r] \text{ rel}_1 \eta(y, \omega_1)[s]) \wedge \text{Intrel}(\eta(x, \omega_2)[r] \text{ rel}_2 \eta(y, \omega_2)[s]) \Rightarrow \text{rel}_1 = \text{rel}_2$ •

*A.2 : Proof of Theorem 3.25*
1. Let MRT(x, y, k) $\land$ 0 ≤ p ≤ length(x) -1 $\land$ 0 ≤ q ≤ length(y) -1 be true for some eventualities x, y, time interval k, and ordinals p, q.
2. Now we will prove the two sides of the equivalence.

*Only if*

3. Suppose r, s exists such that for some interval $\omega$, it is the case that
   $\omega \in \Omega(x, y) \land$ Intrel ($\omega$, within, k) $\land$
   $\neg$Intrel($\phi(x_p, \eta(x, \omega)[r])$, disjoint, $\phi(y_q, \eta(y, \omega)[s])$)
4. In that case, it follows by Definition 3.5 that
   MT($x_p + y_q$, common($\phi(x_p, \eta(x, \omega)[r])$, $\phi(y_q, \eta(y, \omega)[s])$))

5. From the Definition 3.7 of $\phi$ and Axiom 3.9 about $\eta$ it follows that
   Intrel($\phi(x_p, \eta(x, \omega)[r])$, within $\omega$) and
   Intrel($\phi(y_q, \eta(y, \omega)[s])$ within $\omega$)
6. common($\phi(x_p, \eta(x, \omega)[r])$, $\phi(y_q, \eta(y, \omega)[s])$) is within ($\phi(x_p, \eta(x, \omega)[r])$ and
   common($\phi(x_p, \eta(x, \omega)[r])$, $\phi(y_q, \eta(y, \omega)[s])$) is within ($\phi(y_q, \eta(y, \omega)[s])$

7. Intrel ($\omega$, within, k)
8. From 6,5 and 7 and the fact that within is transitive:
   Intrel(common($\phi(x_p, \eta(x, \omega)[r])$, $\phi(y_q, \eta(y, \omega)[s])$, k)
   8 and 4 conclude the proof

*If part*

We will pursue the contra positive of the if part
9. Suppose the RHS of the equivalence is not true, i.e.:
   $\neg \exists$ r, s, $\omega$. $\omega \in \Omega(x, y) \land$ Intrel($\omega$. sub, k) $\land$
   $\neg$( Intrel($\phi(x_p, \eta(x, \omega)[r])$, disjoint, $\phi(y_q, \eta(y, \omega)[s])$))

10. Then it follows by from 9 that
    $\forall$r, s., $\omega$. $\omega \notin \Omega(x, y) \lor \neg$ Intrel($\omega$. sub, k) $\lor$
    Intrel($\phi(x_p, \eta(x, \omega)[r])$, disjoint, $\phi(y_q, \eta(y, \omega)[s])$)

11. It follows from 10 above that:
    $\forall$r, s $\omega$. $\omega \in \Omega(x, y) \land$ Intrel($\omega$. sub, k) $\Rightarrow$
    Intrel($\phi(x_p, \eta(x, \omega)[r])$, disjoint, $\phi(y_q, \eta(y, \omega)[s])$)

12. Let the RHS of 11 hold then by Theorem 3.16:
    ($\omega = k \lor \exists k_1$. cover($\omega, k_1$) = k $\land$ MRT(x, y, $k_1$))

13. Thus From 1 and above it follows that:
    $\exists \omega_1$. $\omega_1 \in \Omega(x, y) \land$ ($\omega_1 = k \lor$
    $\exists k_1$. cover($\omega_1, k_1$) = k $\land$ MRT(x, y, $k_1$))

14. From 13 and Definition 2.2.6 of cover: Intrel($\omega_1$, sub, k)

15. We infer from 11, 13 and 14 above that :
    $\forall$r, s. Intrel($\phi(x_p, \eta(x, \omega_1)[r])$, disjoint, $\phi(y_q, \eta(y, \omega_1)[s])$)

16. We recall by the definition of $\phi$ that $\phi(x_p, \eta(x, \omega_1)[r]$ and $\phi(y_q, \eta(y, \omega_1)[s]$ are the only two sub intervals within $\eta(x, \omega_1)[r]$ and $\eta(y, \omega_1)[s]$ respectively such that xp and yq are maximally true.

    $MT(x_p, \phi(x_p, \eta(x, \omega_1)[r])$ and $MT(y_q, \phi(y_q, \eta(y, \omega_1)[s]))$

17. By Theorem 3.21, $\omega_1 = cover^*(\eta(x, \omega_1))$ and $\omega_1 = cover^*(\eta(y, \omega_1))$
18. It follows then from 15, 16 and 17 that:

    $\neg\exists j, i.\ Intrel(j, within, \omega_1) \wedge Intrel(i, within, \omega_1) \wedge MT(x_p, j) \wedge MT(y_q, i) \wedge$
    $\neg Intrel(j\ disjoint, i)$

19. Thus by Definition 3.5, and 18 above,

    $\neg\exists m.\ Intrel(m, within, \omega_1) \wedge MT(x_p + y_q, m)$

20. From 13 above

    $\omega_1 = k \vee \exists k_1.\ cover(\omega_1, k_1) = k \wedge MRT(x, y, k_1)$

Reasoning by cases:
21. Case 1 : $\omega_1 = k$
22. $\neg\exists m.\ Intrel(m, within, k) \wedge MT(x_p + y_q, m)$ and the contra positive proof is concluded.

23. Case 2: $cover(\omega_1, k_1) = k \wedge MRT(x, y, k_1)$
24. $\neg\exists m.\ Intrel(m, within, \omega_1) \wedge MT(x_p + y_q, m)$
25. From 23 and 24, it follows that:

    $\exists m.\ Intrel(m, within, k) \wedge MT(x_p + y_q, m) \Leftrightarrow$
    $\exists m.\ Intrel(m, within, k_1) \wedge MT(x_p + y_q, m)$

26. From 18. $MRT(x, y, k_1)$
27. By induction on step 1, one concludes that:

    $\neg\exists m.\ Intrel(m, within, k) \wedge MT(x_p + y_q, m)$ and the proof is concluded $\bullet$